# User Intent to Use DeepSeek for Healthcare Purposes and their Trust in the Large Language Model: Multinational Survey Study

Avishek Choudhury, Ph.D., Yeganeh Shahsavar, MSc, and Hamid Shamszare, MSc

***Abstract*—Large language models (LLMs) increasingly serve as interactive healthcare resources, yet user acceptance remains underexplored. This study examines how ease of use, perceived usefulness, trust, and risk perception interact to shape intentions to adopt DeepSeek, an emerging LLM-based platform, for healthcare purposes. A cross-sectional survey of 556 participants from India, the United Kingdom, and the United States was conducted to measure perceptions and usage patterns. Structural equation modeling assessed both direct and indirect effects, including potential quadratic relationships. Results revealed that trust plays a pivotal mediating role: ease of use exerts a significant indirect effect on usage intentions through trust, while perceived usefulness contributes to both trust development and direct adoption. By contrast, risk perception negatively affects usage intent, emphasizing the importance of robust data governance and transparency. Notably, significant non-linear paths were observed for ease of use and risk, indicating threshold or plateau effects. The measurement model demonstrated strong reliability and validity, supported by high composite reliabilities, average variance extracted, and discriminant validity measures. These findings extend technology acceptance and health informatics research by illuminating the multifaceted nature of user adoption in sensitive domains. Stakeholders should invest in trust-building strategies, user-centric design, and risk mitigation measures to encourage sustained and safe uptake of LLMs in healthcare. Future work can employ longitudinal designs or examine culture-specific variables to further clarify how user perceptions evolve over time and across different regulatory environments. Such insights are critical for harnessing AI to enhance outcomes.**

## I. INTRODUCTION

DeepSeek, an open-source large language model (LLM) developed in 2023, has gained attention for its cost-effectiveness and comparable performance to GPT-4. In January 2025, it released a free chatbot application based on DeepSeek-R1, which quickly surpassed ChatGPT as the most-downloaded free app on the U.S. iOS App Store. Its rapid rise and potential to outperform other LLMs can influence public perception and trust. Users may experience a *halo effect*, assuming DeepSeek's competence in all domains due to its coherent and contextually relevant answers. This over-reliance can lead to the abandonment of judgment and cross-checking processes when an AI system seems superior [1]. The fluid and human-like nature of DeepSeek's responses further reinforces the illusion of genuine understanding, blurring the lines between machine output and expert advice. Consequently, users may implicitly trust DeepSeek's answers, even in critical areas like healthcare, without considering potential errors or biases [2, 3].

According to the media equation theory, suggests that people often treat mediated interactions (including those with computers and other digital systems) similarly to real-life interpersonal encounters. According to this view, users might ascribe social qualities to AI systems, causing them to behave as though they are conversing with a trustworthy individual rather than a machine [4]. This tendency is further amplified by an LLM's (like DeepSeek) ability to produce fluid, contextually aware dialogue, leading users to feel comfortable disclosing personal details as they would with a trusted confidant. Its widespread availability in user-friendly formats, like smartphone apps, lowers friction and leads to uncritical acceptance of outputs, increasing privacy risks [5]. Corporate servers store and analyze every query and response, potentially creating a vast repository of personal information. While anonymization measures exist, data breaches and unauthorized re-identification remain concerns, especially without transparent data governance [6, 7].

The AI arms race further complicates data security. DeepSeek must navigate evolving privacy legislation across multiple jurisdictions. Malicious entities may exploit LLMs' authority by designing targeted phishing attacks disguised as official DeepSeek responses. The social credibility of DeepSeek makes it an ideal vehicle for sophisticated fraud

Avishek Choudhury is an assistant professor at West Virginia University, Morgantown, WV, 26506 USA (e-mail: avishek.choudhury@mail.wvu.edu).

Yeganeh Shahsavar is with the Industrial and Management Systems Engineering department at West Virginia University, Morgantown, WV 26506 USA (e-mail: ys00022@mix.wvu.edu).

Hamid Shamszare is with the Industrial and Management Systems Engineering department at West Virginia University, Morgantown, WV 26506 USA (e-mail: hs00055@mix.wvu.edu).

The study, bearing the Institutional Review Board (IRB) protocol number 2302725983 and classified as a flex protocol type, received approval from West Virginia University.

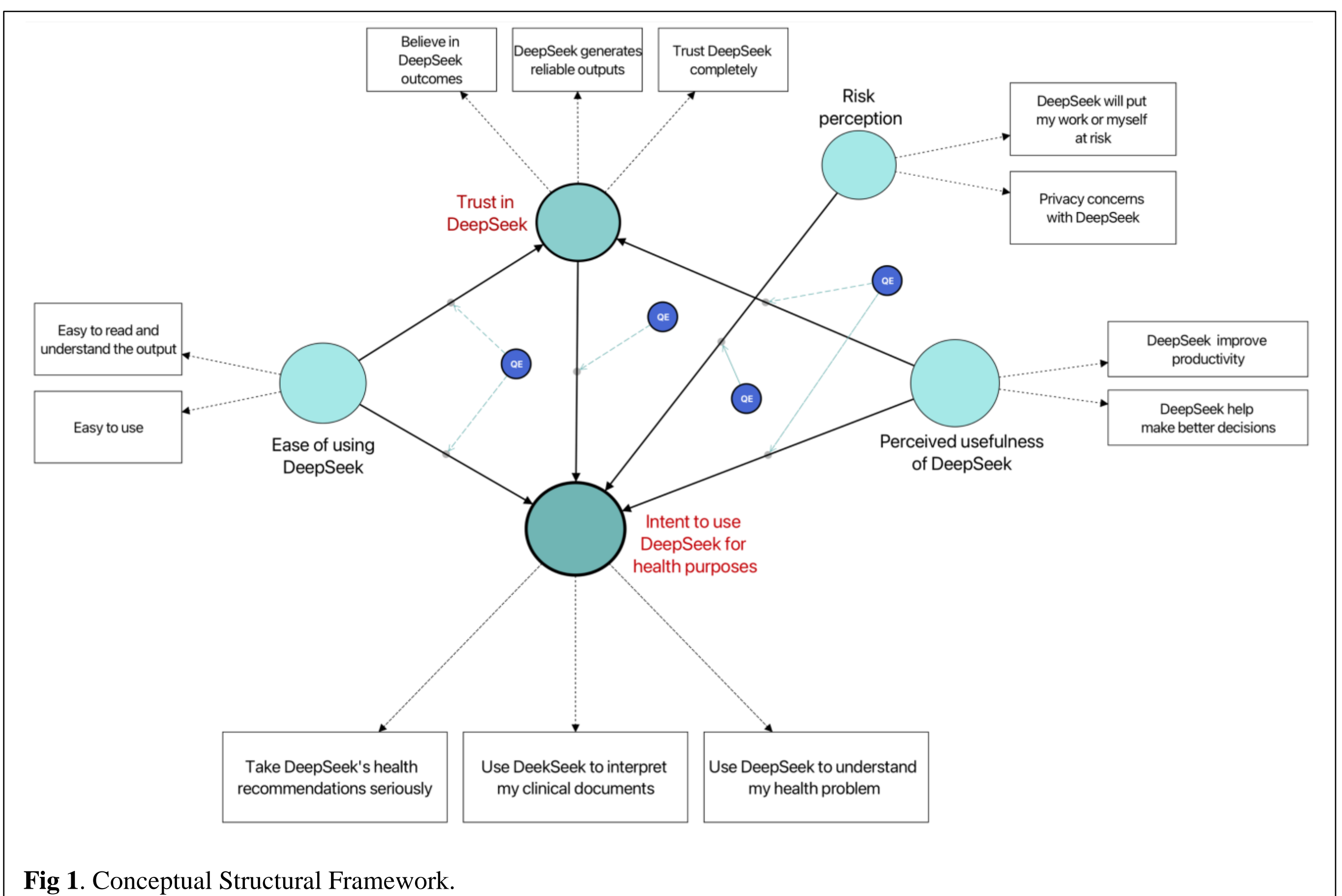

**Fig 1**. Conceptual Structural Framework.

if safeguards are lacking [8]. Even without malicious intent, technical limitations like hallucinations can confuse users and propagate misinformation if the system is trusted unconditionally [9, 10].

DeepSeek's or comparable LLM's potential for biased outputs highlights another concern [11]. Despite cost-effective training on extensive corpora, LLMs reflect biases in their training data. Users may accept DeepSeek's statements as objective truths, leading to unchallenged cultural or social biases. Transparent disclosures about data sources, training methodologies, and error margins are crucial to help users understand when DeepSeek might give skewed answers.

Addressing these issues requires a multidisciplinary effort [12]. While DeepSeek's achievements are groundbreaking, sustaining public trust requires understanding how people perceive and intend to use this technology. Collective awareness and responsibility are essential to harness its benefits without compromising privacy, security, or information integrity.

## II. Research Objective And Hypotheses

The objective of this study is to examine how perceptions of ease of use, risk, and perceived usefulness interact with trust to ultimately shape users' intentions to employ DeepSeek in health-related contexts. In doing so, the research extends established technology acceptance and trust theories by incorporating both direct and indirect paths, as well as by exploring potential non-linear (quadratic) effects [13, 14].

Fig 1 depicts the conceptual model and the hypothesized linkages among these constructs. A key focus centers on the role of ease of use—a construct traditionally linked to technology acceptance. Specifically, the study posits that ease of use will significantly foster trust in DeepSeek and simultaneously increase users' intent to adopt it for health purposes. This expectation aligns with the view that a user-friendly interface reduces cognitive strain, thereby instilling an impression of competence and reliability [15]. Moreover, straightforward navigation of an AI system in a clinical or personal health setting can lower barriers to initial adoption, prompting individuals to incorporate the technology more readily into their decision-making. At the same time, our study acknowledges the possibility of non-linear relationships [16]. That is, once perceived ease of use surpasses a certain threshold, further enhancements may yield diminishing effects on trust or usage intent—hence the inclusion of quadratic testing.

Another central component is trust, which is widely recognized as a pivotal determinant of individuals' willingness to delegate sensitive tasks to AI systems [17]. Here, trust is posited to positively influence intent to use DeepSeek, reflecting the premise that users must believe in the system's credibility and accuracy before relying on its outputs for personally significant choices, such as interpreting medical documents. In parallel, the model predicts that perceived usefulness—the degree to which users believe DeepSeek effectively aids in accomplishing tasks—will both feed into trust and directly motivate the decision to adopt [18]. If DeepSeek demonstrably improves efficiency, offers relevant information, or yields better outcomes, individuals are more likely to view the technology not only as beneficial but also as deserving of confidence, thus reinforcing their intention to use.

In contrast, risk perception is anticipated to exert a negative impact on intent to use [17]. Particularly in health-related scenarios, concerns over data privacy, the possibility of incorrect diagnoses, or broader ethical dilemmas can act as inhibitors to adoption. Users who sense a high level of risk may hesitate to rely on an AI system, even if they recognize certain advantages [19]. The non-linear aspect of this path allows for the possibility that moderate levels of risk perception might exert disproportionately strong effects—potentially deterring adoption more acutely than either very low or very high-risk perceptions.

Additionally, the framework addresses mediation pathways, where trust operates as an intermediary in two distinct relationships. First, the study investigates whether trust mediates the link between ease of use and intent to use, positing that user-friendly design can bolster trust, thereby reinforcing the inclination to adopt DeepSeek. Second, perceived usefulness is hypothesized to enhance trust, which in turn increases intent to use, reflecting the theory that tangible benefits establish an underlying belief in the system's reliability. Both mediating paths also incorporate the possibility that their effects may plateau or intensify under certain conditions—further underscoring the need to examine quadratic relationships. We explore the following eight hypotheses (H):

- H1: Ease of Use positively influences Trust in DeepSeek.
- H2: Ease of Use positively influences Intent to Use DeepSeek for Health-Related Purposes.
- H3: Trust in DeepSeek positively influences Intent to Use DeepSeek for Health-Related Purposes.
- H4: Risk Perception negatively influences Intent to Use DeepSeek for Health-Related Purposes.
- H5: Perceived Usefulness positively influences Trust in DeepSeek.
- H6: Perceived Usefulness positively influences Intent to Use DeepSeek for Health-Related Purposes.
- H7: Trust in DeepSeek mediates the relationship between Ease of Use and Intent to Use DeepSeek.
- H8: Trust in DeepSeek mediates the relationship between Perceived Usefulness and Intent to Use DeepSeek.

## III. Methods

The study, bearing the Institutional Review Board protocol number 2302725983 and classified as a flex protocol type, received approval from West Virginia University.

### A. Survey Instrument

The survey items were adapted from established measurement scales commonly used in technology acceptance and human–computer interaction research [20-22]. A preliminary literature review identified key constructs (e.g., ease of use, trust, risk perception, perceived usefulness, and usage intentions) pertinent to AI-driven applications, particularly in the healthcare domain [17, 23]. Existing, validated items were then modified linguistically to reflect DeepSeek's functionality. For instance, several questions included references to *reliability* and *accuracy*, adapted from trust in automation scales, while risk perception measures were framed to address data privacy and potential adverse outcomes in health-related tasks [24-26].

The resultant questionnaire comprised 12 primary items, each measured on a four-point forced Likert scale (Table I). Questions were grouped to form latent construct and validated. The instrument also had questions about participant demographics. Additionally, the survey incorporated a checking question to verify that respondents thoroughly read all questions before providing their answers, further ensuring data quality.

### B. Pilot Testing and Sampling

Before launching the main data collection, the survey was pilot tested with a small convenience sample (n = 20) who met the criterion of having used DeepSeek at least once in the previous two weeks. Pilot participants were asked to provide open-ended feedback on item clarity, redundancy, and overall length. Minor revisions were made, including refining the wording of risk-related items and adjusting the Likert anchors for consistency.

Following pilot testing, an online version of the final questionnaire was administered via a paid audience paneling service. The survey was distributed to India, United Kingdom (UK), and United States of America (USA). Participant recruitment targeted adult users (18 years or older) who reported having used DeepSeek at least once in the preceding two weeks. Exclusion criteria included individuals with no prior exposure to DeepSeek, as the survey required firsthand experience to accurately assess perceptions of trust, usability, and risk. The questionnaire's introduction reiterated the voluntary nature of participation, guaranteed anonymity, and provided contact details for the principal investigator.

### C. Data Collection Procedure

Data collection took place over a two-week period. Each participant received an individualized survey link, which led to a landing page containing an informed consent statement. After consenting, participants proceeded to the main survey. The survey platform automatically recorded session details, including session ID, IP address (to prevent deduplication only). All identifying information was removed prior to data analysis. Participants could terminate the survey at any point without penalty.

TABLE I
THE SURVEY INSTRUMENT

<table>
<tr><th>Latent Construct</th><th>Survey Questions</th></tr>
<tr><td rowspan="3">Trust in DeepSeek →</td><td>To what extent do you agree or disagree with the following: I would trust DeepSeek completely. (T1)</td></tr>
<tr><td>To what extent do you agree or disagree with the following: DeepSeek generates reliable outputs. (T2)</td></tr>
<tr><td>To what extent do you agree or disagree with the following: I believe in DeepSeek outcomes. (T3)</td></tr>
<tr><td rowspan="3">Intent to use DeepSeek for Health Purposes →</td><td>To what extent do you agree or disagree with the following: I would take DeepSeek's health related recommendations seriously. (IU1)</td></tr>
<tr><td>To what extent do you agree or disagree with the following: I will use DeepSeek to interpret my clinical documents. (IU2)</td></tr>
<tr><td>To what extent do you agree or disagree with the following: I want to use DeepSeek to understand minor health problem I occasionally get. (IU3)</td></tr>
<tr><td rowspan="2">Ease of Using DeepSeek →</td><td>To what extent do you agree or disagree with the following: I find it easy to read and understand the output generated by DeepSeek. (E1)</td></tr>
<tr><td>To what extent do you agree or disagree with the following: | DeepSeek is easy to use. (E2)</td></tr>
<tr><td rowspan="2">Perceived Usefulness of DeepSeek →</td><td>To what extent do you agree or disagree with the following: DeepSeek generated outcomes will help me improve productivity. (PU1)</td></tr>
<tr><td>To what extent do you agree or disagree with the following: DeepSeek generated outcomes will help me make better decisions. (PU2)</td></tr>
<tr><td rowspan="2">Risk Perception →</td><td>To what extent do you agree or disagree with the following: | Using DeepSeek will put my work or myself at risk. (R1)</td></tr>
<tr><td>To what extent do you agree or disagree with the following: | I have privacy concerns with DeepSeek. (R2)</td></tr>
<tr><td colspan="2">Note: Risk Perception was reverse coded before performing any analyses</td></tr>
</table>

### D. Statistical Analyses

First, descriptive statistics were calculated for all survey questions to provide an overview of the responses' central tendency, dispersion, and distribution. These statistics offered initial insights into the participants' attitudes and perceptions regarding the constructs under investigation. Following the descriptive analysis, the research team employed Partial Least Squared Structural Equation Modeling (PLS SEM) to examine the relationships between the latent constructs. PLS-SEM is a multivariate technique that allows estimation of complex cause-effect relationships between latent constructs and their indicators [27]. This method was chosen for its ability to handle small to medium-sized samples and suitability for exploratory research [28].

The PLS-SEM analysis in our study was conducted in two stages: the assessment of the measurement model and the evaluation of the structural model. We assessed the measurement model for reliability and validity by focusing on four aspects: indicator reliability, internal consistency reliability, convergent validity, and discriminant validity. Indicator reliability was examined by analyzing the factor loadings of each indicator, with loadings greater than 0.5 considered satisfactory. We evaluated internal consistency reliability using composite reliability (rhoC) [29]. Convergent validity was assessed by examining the average variance extracted (AVE), and values above 0.5 indicated an adequate convergent validity [29, 30]. In addition to these assessments, we also evaluated the reliability of the constructs in our research model using the average inter-item correlation (rho_c). A value of 0.7 or higher for rhoC is generally considered to indicate satisfactory reliability.

After confirming the measurement model's adequacy, we evaluated the structural model to test our hypotheses. We performed a bootstrap resampling procedure with 10,000 iterations to obtain the parameter estimates and compute confidence intervals.

### E. Sample Size Justification

Structural equation modeling commonly benefits from relatively large sample sizes to ensure robust parameter estimation and adequate statistical power. Guidelines in the SEM literature vary, but several rules of thumb have been proposed where some recommends a minimum of 200 participants for most SEM applications, while others suggest having at least five to ten respondents per estimated parameter in the model [31-33].

In our study, the final instrument contained 12 key items spanning multiple latent constructs (e.g., ease of use, trust, risk perception, perceived usefulness, and intent to use). Consequently, typical rules of thumb would indicate a desired sample size in the range of 300–500 participants to comfortably meet assumptions of stable parameter estimation.

## III. RESULTS

A total of 556 complete responses were collected. Geographically, 184 individuals were based in India, 185 in the UK, and 187 in the United States, reflecting an almost even split across these three locales. Of the 556 participants in this study, 33% used it "once a month," 28% used it "once a week," 25% used it "more than once per week," and 14% used it "almost every day." Regarding education, most respondents (38%) held a bachelor's degree, followed by master's degree holders (28%), high school graduates (26%), those with some high school (4%), and doctoral-level degrees (4%). The sex distribution was nearly balanced, with 49% identifying as male and 51% as female. In terms of age, the largest proportion of respondents fell in the 26–35-year range (25%), while 14% were aged 18–25, 19% were 36–45, another 19% were 46–55, 15% were 56–65, and 7% were 66 or older.

TABLE II
THE MEASUREMENT MODEL AND VALIDATION

| Observed Variable | Constructs | Factor Loading | AVE | Rho_c | Cronbach's alpha |
|---|---|---|---|---|---|
| *T1* → | Trust in DeepSeek | 0.82 | 0.65 | 0.85 | 0.73 |
| *T2* → | | 0.78 | | | |
| *T3* → | | 0.82 | | | |
| *IU1* → | Intent to use DeepSeek for Health Purposes | 0.85 | 0.70 | 0.88 | 0.79 |
| *IU2* → | | 0.84 | | | |
| *IU3* → | | 0.82 | | | |
| *E1* → | Ease of Using DeepSeek | 0.90 | 0.82 | 0.90 | 0.78 |
| *E2* → | | 0.91 | | | |
| *PU1* → | Perceived Usefulness of DeepSeek | 0.86 | 0.77 | 0.87 | 0.71 |
| *PU2* → | | 0.90 | | | |
| *R1* → | Risk Perception | 0.96 | 0.66 | 0.79 | 0.70 |
| *R2* → | | 0.62 | | | |

## A. Measurement Model and Survey Validation

Table II indicate a robust measurement model with strong evidence of convergent and discriminant validity across all latent constructs. Convergent validity is supported by high AVE values: each construct exceeds the commonly accepted 0.50 threshold (ranging from 0.650 to 0.820), indicating that the items explain a substantial portion of their respective constructs. Likewise, rho_c and Cronbach's alpha values are mostly well above the 0.70 benchmark confirming internal consistency among indicators.

Discriminant validity is supported by the Heterotrait–Monotrait (HTMT) ratios, nearly all of which fall below the critical value of 0.90 or 1.00. In addition, the $R^2$ values of 0.532 (adjusted 0.525) for Intent to Use DeepSeek for Health Purposes and 0.592 (adjusted 0.589) for Trust in DeepSeek indicate that the explanatory power of the proposed paths is moderate to strong. The significant T-statistics and p-values across the key relationships further verify that the conceptualized constructs have been measured accurately, with only a few non-significant quadratic paths. These findings validate the measurement instrument and confirm that the theorized latent variables—Ease of Using DeepSeek, Perceived Usefulness, Trust in DeepSeek, Risk Perception, and Intent to Use DeepSeek for Health Purposes—exhibit sufficient reliability and validity to warrant confidence in subsequent structural analyses.

In addition to the overall validity of the measurement model, the findings also support the validity of the individual survey questions that served as indicators for each latent construct. First, the high factor loadings (as implied by the satisfactory AVE values) suggest that each item meaningfully contributes to measuring its intended factor—i.e., users' responses to questions about ease of use, trust, perceived usefulness, risk, or intent to use strongly correlate with the respective latent constructs they were designed to represent. Second, the internal consistency indices (Cronbach's alpha and composite reliability) confirm that groups of questions intended to measure the same construct hang together well, indicating the survey items reliably capture the same underlying concept. Finally, the discriminant validity checks (heterotrait monotrait ratios) confirm that sets of questions targeting one construct do not overlap excessively with those measuring other constructs; this indicates that the items' wording and content domains effectively capture distinct dimensions of user perceptions toward DeepSeek. Taken together, these indicators demonstrate that the survey questions—and not just the overarching constructs—exhibit satisfactory validity.

## B. Structural Equation Model (Direct, Indirect, and Total Effects)

As shown in Table III, the path coefficients and mediation analyses provide strong evidence for most of the hypothesized relationships, though one notable exception emerged regarding risk perception.

*Ease of Use (H1)* exhibits a significantly positive direct effect on Trust in DeepSeek, indicating that when users perceive DeepSeek to be more intuitive and less effortful to operate, their trust in the system increases correspondingly.

However, the direct relationship from *Ease of Use* to *Intent to Use DeepSeek for health-related purposes (H2)* is not statistically significant. Despite this, a significant total effect and a strong indirect path through *Trust* confirm that trust fully mediates the impact of ease of use on user intentions. Hence, while there is no direct link from ease of use to intent, ease of use indirectly drives intent via trust, satisfying *H2's* broader premise when mediation is considered.

TABLE III
THE STRUCTURAL MODEL

| Paths | Direct effect | Indirect effect | Total effect |
|---|---|---|---|
| | β (Std. Dev) | | |
| Ease of Using DeepSeek → Intent to Use DeepSeek for Health Purposes | 0.07 (0.058) | | 0.23 (0.06)*** |
| Ease of Using DeepSeek → Trust in DeepSeek | 0.36 (0.05)*** | | 0.36 (0.05)*** |
| Perceived Usefulness of DeepSeek → Intent to Use DeepSeek for Health Purposes | 0.17 (0.05)*** | | 0.40 (0.04)*** |
| Perceived Usefulness of DeepSeek → Trust in DeepSeek | 0.52 (0.04)*** | | 0.52 (0.04)*** |
| Risk Perception → Intent to Use DeepSeek for Health Purposes | 0.20 (0.03)*** | | 0.20 (0.03)*** |
| Trust in DeepSeek → Intent to Use DeepSeek for Health Purposes | 0.45 (0.05)*** | | 0.45 (0.05)*** |
| Ease of Using DeepSeek → Trust in DeepSeek → Intent to Use DeepSeek for Health Purposes | | 0.16 (0.02)*** | 0.16 (0.02)*** |
| Perceived Usefulness of DeepSeek → Trust in DeepSeek → → Intent to Use DeepSeek for Health Purposes | | 0.23 (0.03*** | 0.23 (0.03)*** |
| β=standardized coefficient; ***= *p<0.001* | | | |

*Trust in DeepSeek (H3)* exerts a strong, positive influence on *Intent to Use DeepSeek*, suggesting that once users develop confidence in DeepSeek's outputs, they are more inclined to rely on it for health-focused tasks.

*Risk Perception (H4)* shows a significant negative path coefficient, supporting the hypothesized relationship. This implies that higher perceived risk—as captured in the current measurement—correlates with a lower intent to use DeepSeek, deterring adoption.

The findings strongly support both direct and mediated effects concerning Perceived Usefulness. *Perceived Usefulness (H5)* significantly increases *Trust in DeepSeek*, indicating that when participants see value in DeepSeek's capabilities, their faith in the technology grows.

*Perceived Usefulness* also shows a robust direct effect on *Intent to Use DeepSeek (H6),* underscoring the salience of benefit driven evaluations in motivating adoption. Additionally, mediation analyses reveal that trust partially mediates the effect of perceived usefulness on intent: the direct path remains significant, while the indirect path further boosts intentions.

Finally, *H7* and *H8* both concern mediation by trust. For *H7 (Ease of Use → Trust → Intent)*, the indirect path is significant while the direct path is not, signifying a full mediation scenario in which ease of use shapes intentions solely through trust. For *H8 (Perceived Usefulness → Trust → Intent)*, there is evidence of partial mediation, as both the direct and indirect paths are significant.

## B. Structural Equation Model (Quadratic Effects)

The model also explored potential quadratic (QE) non-linear relationships among the key constructs. The results indicate that the quadratic effect of *Ease of Use* on *Intent to Use DeepSeek for Health Purposes* is significant ($\beta = 0.072$, $p = 0.011$), suggesting that beyond a certain point, the influence of ease of use on user intentions may intensify or plateau rather than progressing linearly. In contrast, the quadratic term of Ease of Use on Trust in DeepSeek was not significant ($\beta = 0.038$, $p = 0.102$), indicating that trust levels respond to ease of use in a linear fashion.

The quadratic effect of *Risk Perception* on *Intent to Use DeepSeek for Health Purposes* emerged as significant ($\beta = -0.119$, $p = 0.002$), implying that moderate risk perceptions might not affect adoption the same way extremely low or very high-risk perceptions do—potentially reflecting a threshold.

Perceived Usefulness exhibited a negative but significant quadratic relationship with Intent to Use ($\beta = -0.074$, $p = 0.045$), suggesting that at very high levels of perceived usefulness, additional increases might yield diminishing or even marginally negative returns on user intentions. By contrast, no significant quadratic effect of Perceived Usefulness was detected on Trust in DeepSeek ($\beta = 0.022$, $p = 0.439$), indicating that trust is likely more directly related to usefulness without an apparent non-linear inflection point.

Collectively, these results demonstrate that while ease of use, risk perception, and perceived usefulness can shape user intentions in curvilinear ways, not all relationships in the model exhibit non-linear patterns. For constructs like trust, effects appear to follow a more straightforward linear trajectory. These findings underscore the importance of accounting for potential threshold or plateau effects in user adoption research, particularly in health-oriented applications where attitudes and behaviors can shift dramatically once certain levels of risk or perceived benefits are reached. Figure 3 summarizes the overall findings.

# IV. DISCUSSION

This study makes a significant contribution to the field of informatics by systematically examining user interactions with DeepSeek, a novel AI platform, through the lens of data governance, privacy, and security. The structured measurement model employed in this study provides valuable insights into how users perceive and respond to DeepSeek's capabilities. By capturing and analyzing users' perceptions of risk, trust, and ease of use, the study addresses critical informatics issues such as clinical decision support and system design. In addition to its focus on informatics, this work adopts a human factors and technology acceptance perspective. While previous research has explored these constructs in the context of other large language models (e.g., ChatGPT), this study is among the first to empirically validate a comprehensive measurement model specifically tailored for DeepSeek. This model integrates both direct and non-linear effects, demonstrating that user perceptions of system utility and trustworthiness are paramount in health-oriented AI contexts. The interplay between ease of use, perceived usefulness, trust, and risk perception provides a comprehensive framework for understanding user intentions to adopt DeepSeek. This highlights the importance of developers prioritizing these factors to enhance user acceptance.

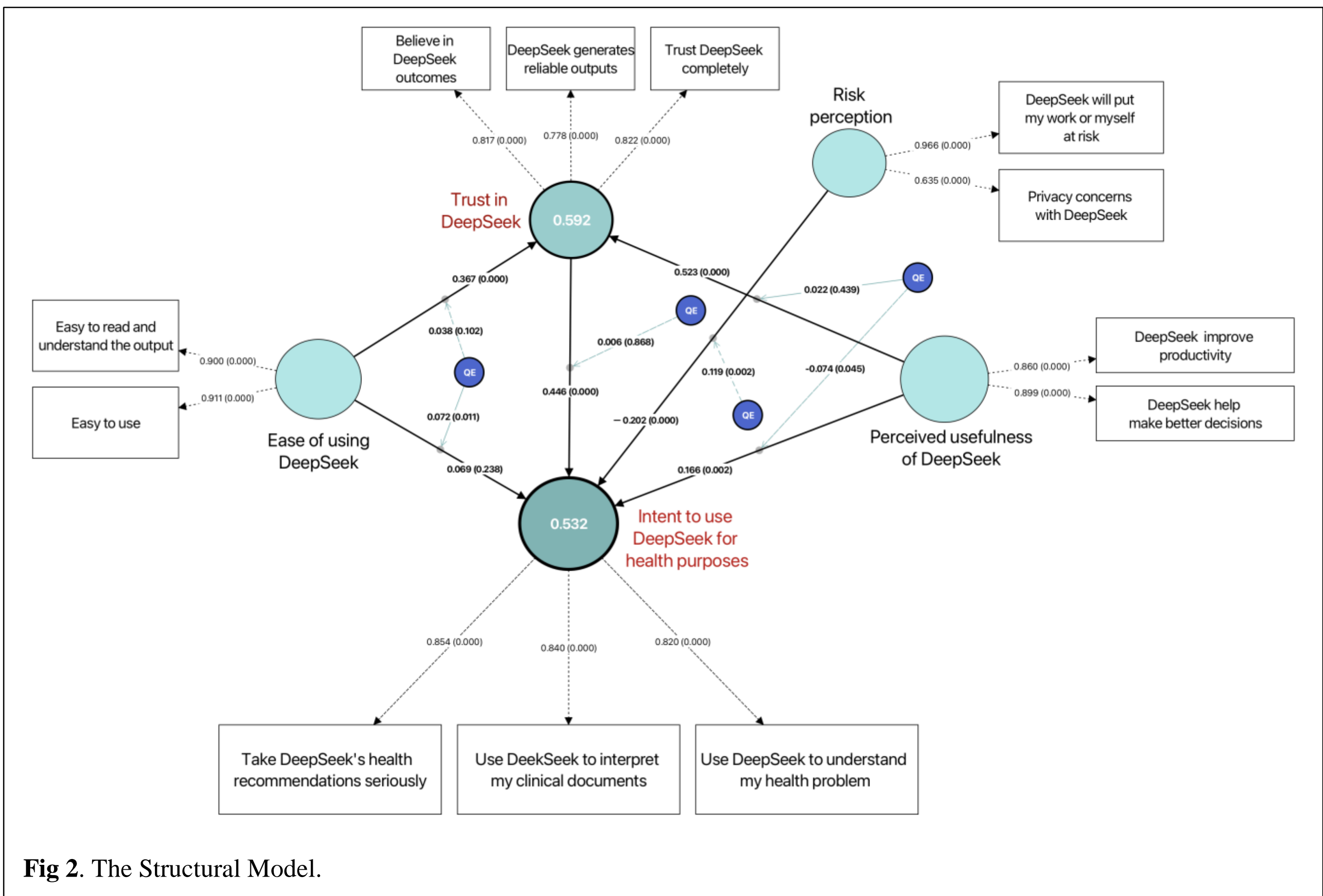

**Fig 2**. The Structural Model.

The most notable contribution of this study lies in its comprehensive examination of how ease of use, perceived usefulness, trust, and risk perception influence users' intentions to adopt DeepSeek for health-related purposes. By drawing on a large sample of 556 participants—fairly evenly distributed across India, the UK, and the United States—this research demonstrates that *trust* is a primary determinant of users' adoption intentions, mediating the relationships from both ease of use and perceived usefulness to intent. This aligns with existing literature that emphasizes the critical role of trust in the acceptance of AI technologies in healthcare settings. For instance, a highlights that lack of trust in AI systems is a significant barrier to their adoption in medical care, indicating that enhancing user confidence is essential for successful implementation [34].

While ease of use does not directly predict intent to use (H2), it indirectly exerts a strong effect via trust (H7)—highlighting that enhancing usability alone is insufficient unless it also fosters greater confidence in the system. This notion is echoed in the work of Schnall et al., who argue that both perceived usefulness and ease of use are critical for the successful adoption of mobile health technologies [35].

Furthermore, perceived usefulness drives both trust and intent, highlighting the central role of benefit-driven evaluations in motivating user behavior (H5, H6, H8). These results extend existing technology acceptance theories by showing that user perceptions of system utility and trustworthiness take precedence in health-oriented AI contexts. It also aligns with the findings reported in a meta-analysis demonstrating that perceived benefits significantly influence health-related behaviors [36]. In the context of AI in healthcare, a study found that stakeholders prioritize the utility of AI tools, which supports our assertion regarding the importance of perceived usefulness in driving adoption intentions [37].

Additionally, the negative direct relationship from risk perception to intent to use (H4) supports longstanding expectations that higher apprehension can discourage adoption. In line with our findings a study acknowledged risk perception as a key predictor of preventive behavioral intentions among healthcare workers, reinforcing the notion that risk perceptions play a crucial role in shaping user intentions [38]. However, our findings also suggest that this relationship may manifest non-linearly under certain conditions, which is a nuanced perspective that warrants further exploration in future research.

Although most hypothesized paths proved significant, ease of use did not directly correlate with intent to use (H2), highlighting that simpler interfaces alone might not automatically drive adoption for sensitive applications like health inquiries. Furthermore, the quadratic terms for several relationships were non-significant, suggesting a primarily linear dynamic between these constructs. Such null findings can be instructive: they indicate that, in certain relationships, user attitudes do not exhibit threshold or plateau effects and are instead driven by more consistent incremental changes (e.g., each increase in usefulness incrementally raising trust without a significant non-linear inflection).

Policymakers focused on regulating AI in healthcare can draw from these results by encouraging transparency initiatives that bolster user trust—for instance, mandating disclosures about how AI systems generate medical suggestions or interpret user data. For healthcare providers, the evidence that ease of use indirectly shapes intent via trust means that training programs and user onboarding should emphasize not only user-friendly design but also trust-building strategies (e.g., illustrating system accuracy, addressing privacy concerns). Given that perceived usefulness directly drives adoption, developers should prioritize demonstrating tangible benefits (e.g., better medical outcomes, efficient data processing) to foster acceptance. Lastly, the significant quadratic effects for risk perception and ease of use highlight the need to monitor user attitudes for potential *tipping points* beyond which additional enhancements in usability or risk mitigation have disproportionate impacts on willingness to adopt.

Few constraints must be acknowledged when interpreting these findings. First, this cross-sectional study cannot establish causal direction definitively, although the robust model fit and theoretical grounding strengthen causal inferences. Second, although the sample size was large and relatively balanced across three regions, the results may not generalize to other cultural or regulatory contexts where attitudes toward AI-driven health tools differ. Third, the study relied on self-reported measures, which may be subject to social desirability bias or inaccuracies in recalling frequency and manner of use. Finally, while the model covered central factors in technology acceptance (ease of use, usefulness, trust, risk), there may be other contextual variables—such as domain expertise, quality of healthcare infrastructure, or cultural norms—that also shape AI adoption in health contexts. Future studies could integrate longitudinal or experimental designs to verify these relationships over time and examine how usage behaviors evolve with extended AI exposure in real clinical or personal health settings.